\documentclass[sigconf]{acmart}

\usepackage{booktabs} 
\usepackage{syntax}
\usepackage{listings}
\usepackage{subcaption}
\colorlet{punct}{black!60!black}
\definecolor{background}{HTML}{FFFFFF}
\definecolor{delim}{RGB}{0,0,0}
\colorlet{numb}{black!60!black}
\lstdefinelanguage{json}{
    basicstyle=\normalfont\ttfamily,
    numberstyle=\scriptsize,
    stepnumber=1,
    numbersep=8pt,
    showstringspaces=false,
    breaklines=true,
    frame=lines,
    backgroundcolor=\color{background},
    literate=
     *{0}{{{\color{numb}0}}}{1}
      {1}{{{\color{numb}1}}}{1}
      {2}{{{\color{numb}2}}}{1}
      {3}{{{\color{numb}3}}}{1}
      {4}{{{\color{numb}4}}}{1}
      {5}{{{\color{numb}5}}}{1}
      {6}{{{\color{numb}6}}}{1}
      {7}{{{\color{numb}7}}}{1}
      {8}{{{\color{numb}8}}}{1}
      {9}{{{\color{numb}9}}}{1}
      {:}{{{\color{punct}{:}}}}{1}
      {,}{{{\color{punct}{,}}}}{1}
      {\{}{{{\color{delim}{\{}}}}{1}
      {\}}{{{\color{delim}{\}}}}}{1}
      {[}{{{\color{delim}{[}}}}{1}
      {]}{{{\color{delim}{]}}}}{1},
}

\begin{document}
\title{Talakat: Bullet Hell Generation through Constrained Map-Elites}


\author{Ahmed Khalifa}
\affiliation{%
  \institution{New York University}
  \streetaddress{5 MetroTech Center}
  \city{New York City}
  \state{New York}
}
\email{ahmed.khalifa@nyu.edu}

\author{Scott Lee}
\affiliation{%
  \institution{New York University}
  \streetaddress{5 MetroTech Center}
  \city{New York City}
  \state{New York}
}
\email{sl3998@nyu.edu}

\author{Andy Nealen}
\affiliation{%
  \institution{New York University}
  \streetaddress{5 MetroTech Center}
  \city{New York City}
  \country{New York}}
\email{andy@nealen.net}

\author{Julian Togelius}
\affiliation{%
  \institution{New York University}
  \streetaddress{5 MetroTech Center}
  \city{New York City}
  \country{New York}}
\email{julian@togelius.com}

\renewcommand{\shortauthors}{Khalifa et al.}
%
%

\begin{abstract}
We describe a search-based approach to generating new levels for bullet hell games, which are action games characterized by and requiring avoidance of a very large amount of projectiles. Levels are represented using a domain-specific description language, and search in the space defined by this language is performed by a novel variant of the Map-Elites algorithm which incorporates a feasible-infeasible approach to constraint satisfaction. Simulation-based evaluation is used to gauge the fitness of levels, using an agent based on best-first search. The performance of the agent can be tuned according to the two dimensions of strategy and dexterity, making it possible to search for level configurations that require a specific combination of both. As far as we know, this paper describes the first generator for this game genre, and includes several algorithmic innovations.
%
%
%
\end{abstract}

%
%
\copyrightyear{2018} 
\acmYear{2018} 
\setcopyright{acmcopyright}
\acmConference[GECCO '18]{Genetic and Evolutionary Computation Conference}{July 15--19, 2018}{Kyoto, Japan}
\acmBooktitle{GECCO '18: Genetic and Evolutionary Computation Conference, July 15--19, 2018, Kyoto, Japan}
\acmPrice{15.00}
\acmDOI{10.1145/3205455.3205470}
\acmISBN{978-1-4503-5618-3/18/07}



\maketitle

\section{Introduction}

Games of the same series or genre often share a number of gameplay elements and can sometimes feel as if they play very similarly. As such, games sometimes are differentiated 
based on the quality and quantity of their content. Some genres and series exhibit this emphasis on content more strongly than others, and it is in this space that Procedural Content Generation (PCG) ~\cite{shaker2016procedural} can be particularly useful.

Content in the bullet hell genre is often measured by its challenge, so developers often attempt to make their games punishingly hard. However, players come with a wide variety of skill levels, and so many bullet hell games add multiple difficulty levels in the hope that the game can present challenging content to a wider variety of players. It is entirely possible for designers to miss the mark on easier content, making it too close in difficulty to harder content or making it too easy to be interesting. This is a fundamental challenge that can arise in game design, and one that may be lightened with PCG techniques.

In this paper, we describe a PCG algorithm implementation designed to generate bullet patterns for bullet hell games. Of particular interest to this work is the infrastructure enabling the algorithm to generate the patterns, as well as the algorithm itself, and the means by which created patterns are evaluated. To that end, we present Talakat, a description language designed to encapsulate and describe bullet hell patterns, a variation upon the MAP-Elites algorithm that generates Talakat descriptions, and a simulation evaluation method that guides evolution toward levels of specific challenge along the two dimensions strategy and dexterity.
\section{Background}

Bullet hell is a subgenre of shoot 'em up games, where player characters fire projectiles at enemies, which similarly fire projectiles at the player. The goal of these games is generally to defeat enemies while avoiding the projectiles that are fired at them. Notable shoot 'em ups include \emph{SpaceWar} (CoCoPaPa Soft, 1962) and \emph{Space Invaders} (Taito, 1978). Bullet hell games share this fundamental concept, but are distinguished by a much higher quantity of bullets as well as higher difficulty. Notable recent bullet hell games include \emph{Jamestown: Legend of the Lost Colony} (Final Form Games, 2011) and \emph{Touhou Tenkuushou ~ Hidden Star in Four Seasons} (Team Shanghai Alice, 2017).

The games of the bullet hell genre typically share a single core formula ~\cite{collman2014maku}, and the gameplay of many modern bullet hell games is strikingly similar to that of games from 1995. Instead, many bullet hell games are differentiated by the quality of their levels. This does, however, raise the question of what qualities are desirable in a bullet hell level. Principal among these is playability. Games that are difficult by design often walk a fine line between satisfyingly challenging and simply impossible to complete. Playtesting is of course important in ensuring playability, but a game that is challenging by design will require highly skilled (and harder to find) players to validate a game. 
Additionally, players often expect bullet hell games to be challenging. Some levels heavily emphasize what is referred to as \textit{skillful dodging}, which reward high dexterity and fast reflexes. Bullet hell levels can also require the identification of safe spots or planned paths, and reward intelligent planning and long-term strategy.  

\subsection{Procedural Content Generation}
Both industry and research groups have developed methods of procedural content generation in games. In the industry, procedural level generation has been particularly prevalent in strategy, dungeon-crawling, and role-playing games. PCG in games dates as far back as ~\emph{Rogue} (Epyx, 1980), and remains a mainstay of modern game design. Franchises such as ~\emph{Diablo} (Blizzard, 1996), ~\emph{Mystery Dungeon} (Spike Chunsoft, 1993), and ~\emph{Disgaea} (Nippon Ichi Software, 2003) have featured level generation as a core feature and a key selling point. The idea of effectively infinite content is an appealing one to both consumers and developers.  

Researchers have developed and presented a number of methods that can be used to generate levels in games. Especially popular in this area of research is generation through evolutionary strategies. Shaker et al discuss a means of generating personalized content for Super Mario Bros through grammatical evolution ~\cite{shaker2012evolving}. Browne et al. present the application of genetic programming toward the procedural generation of games through the Ludi system, resulting in the development of games such as \emph{Yavalath} ~\cite{browne2010evolutionary}. Sentient Sketchbook utilizes a user-driven evolutionary algorithm to generate levels for tile-based games ~\cite{liapis2013sentient}. Closer to the generation of bullet patterns is the work presented by Hastings et al., which utilized online user-driven neuroevolution techniques to procedurally generate novel shot types for the game \emph{Galactic Arms Race} ~\cite{hastings2009evolving}. We aim to use a variation on MAP-Elites ~\cite{mouret2015illuminating} with a simulation-based evaluation to measure the quality and difficulty of a created level.

A key requirement of most procedural level generation is a reasonable and workable representation of generated content. For example, the Video Game Description Language (VGDL) represents content using a high-level description language that defines entities and the behaviors between them ~\cite{ebner2013towards}~\cite{schaul2013video}. Game generation using VGDL has involved methodologies that mutate interactions between game entities by modifying the script that generates them ~\cite{nielsen2015towards}. Work involving PCG in Super Mario Bros demonstrates the number of ways a level representation can take. For example, Snodgras et al. generates levels using a higher order Markov chain ~\cite{snodgrass2014experiments}. By contrast, Shaker et al. represented Mario levels using abstract grammar based representations ~\cite{shaker2012evolving}. Tracery also presents a grammar that can be used to procedurally generate game text. The grammar acts as an abstract representation of parameters along which the text is to be generated ~\cite{compton2014tracery}. Using a grammar representation, it is possible to procedurally generate content using techniques akin to grammatical evolution ~\cite{o2003grammatical}.

\subsection{Simulation-based content evaluation}
It is also important to have means to validate and evaluate the generated content. One way to evaluate game content is simulation-based evaluation using AI agents. Smith et al. make use of AI agents to validate the playability of a platformer level in the Tanagra framework ~\cite{smith2011tanagra}. Isaksen et al. use large-scale simulation-based evaluation to explore the space of Flappy Bird variants and identify sets of variables that result in interesting variants ~\cite{isaksen2017exploring}. Silva et al. also demonstrated the usage of AI-based playtesting in gameplay analysis as well as the identification of unexpected imperfections in the \emph{Ticket To Ride} (Days of Wonder, 2004) board game ~\cite{de2017ai}. For bullet hells, one of the most important considerations is difficulty. Difficulty, however, is a complex and fraught topic. Jennings-Teats et al. present an implementation of Dynamic Difficulty Adjustment (DDA) that generates levels with a difficulty specifically tailored toward an individual player, creating harder content as the player improves. This makes use of a feedback-based model, in which player involvment and evaluation is required for the generation of difficult content ~\cite{jennings2010polymorph}. Isaksen et al. present a model of difficulty as a function of dexterity and strategy, as well as an AI-based approach to measuring these quantities ~\cite{isaksen2017simulating}. Bullet hell patterns generally fall neatly within this framework, emphasizing either dexterity-in the form of surgical movements or quick reactions-or strategy-in the form of safe spots, paths, and misdirections.

\section{Talakat}

\begin{figure}
\hrule
\begin{grammar}

<script> ::= <spawners> <boss>

<spawners> ::= <spawner> | <spawner> <spawners>

<spawner> ::= id <spawnerParameter> <spawnedParameter> <bulletParameter>

<spawnerParameter> ::= <spawnPattern> patternTime patternRepeat <angleCSV> <radiusCSV>

<spawnPattern> ::= `bullet' <spawnPattern> | `wait' <spawnPattern> | id <spawnPattern> | $\epsilon$

<spawnedParameter> ::= <numberCSV> <angleCSV> <speedCSV>

<bulletParameter> ::= <radiusCSV> <colorCSV>

<numberCSV> ::= minValue maxValue rate interval <type>

<angleCSV> ::= minValue maxValue rate interval <type>

<speedCSV> ::= minValue maxValue rate interval <type>

<radiusCSV> ::= minValue maxValue rate interval <type>

<colorCSV> ::= minValue maxValue rate interval <type>

<type> ::= `circle' | `inverse'

<boss> ::= bossPosition bossHealth <script>

<script> ::= <scriptEvent> | <scriptEvent> <script>

<scriptEvent> ::= trigger <events>

<events> ::= <event> | <event> <events>

<event> ::= <spawnEvent> | <clearEvent>

<spawnEvent> ::= `spawn' id speed angle | `spawn' `bullet' speed angle

<clearEvent> ::= `clear' id | `clear' `bullets' | `clear' `spawners'

\end{grammar}
\hrule
\caption{\emph{Talakat} language as a context free grammar. Angular brackets values such as <spawners> are non terminal, quoted values such as `bullet' are string terminals, while other values such as minValue are number terminals.}
\label{ex:talakatGrammar}
\end{figure}

Talakat~\footnote{Detailed documentation of the grammar can be found at https://github.com/amidos2006/Talakat/wiki/Scripting-Language} is a description language that describes bullet hell levels. A Talakat script constitutes a single bullet hell level. Figure \ref{ex:talakatGrammar} shows the full grammar of Talakat. Figure \ref{ex:talakatExample} shows an example of a Talakat script. A single script is divided into two parts: the Spawner section and the Boss section. 

\begin{figure}
\begin{lstlisting}[language=json]
{
  spawners:{
    one:{
      pattern:["two"],
      patternTime:"4",
      spawnerAngle:"0,360,10,12,circle",
      spawnedSpeed:"0",
      spawnedNumber:"4",
      spawnedAngle:"360"
    },
    two:{
      pattern:["bullet"],
      patternRepeat:"1",
      spawnedAngle:"30",
      spawnedNumber:"3",
      spawnedSpeed: "4"
    }
    three:{
      pattern:["bullet"],
      patternTime:"4",
      spawnerAngle:"0,180,2,0,reverse",
      spawnedSpeed:"2",
      spawnedNumber:"2",
      spawnedAngle:"360"
    }
  },
  boss:{
    bossHealth: 3000,
    bossPosition: "0.5, 0.2",
    script:[
      {
        health:1,
        events:["spawn,one"]
      },
      {
        health:0.5,
        events:["clear,spawners", "spawn,three"]
      }
    ]
  }
}            
\end{lstlisting}
\caption{An example of a full Talakat script.}
\label{ex:talakatExample}
\end{figure}

\subsection{Spawners Section}


\begin{figure}
\begin{subfigure}{.5\linewidth}
  \centering
  \includegraphics[width=.9\linewidth]{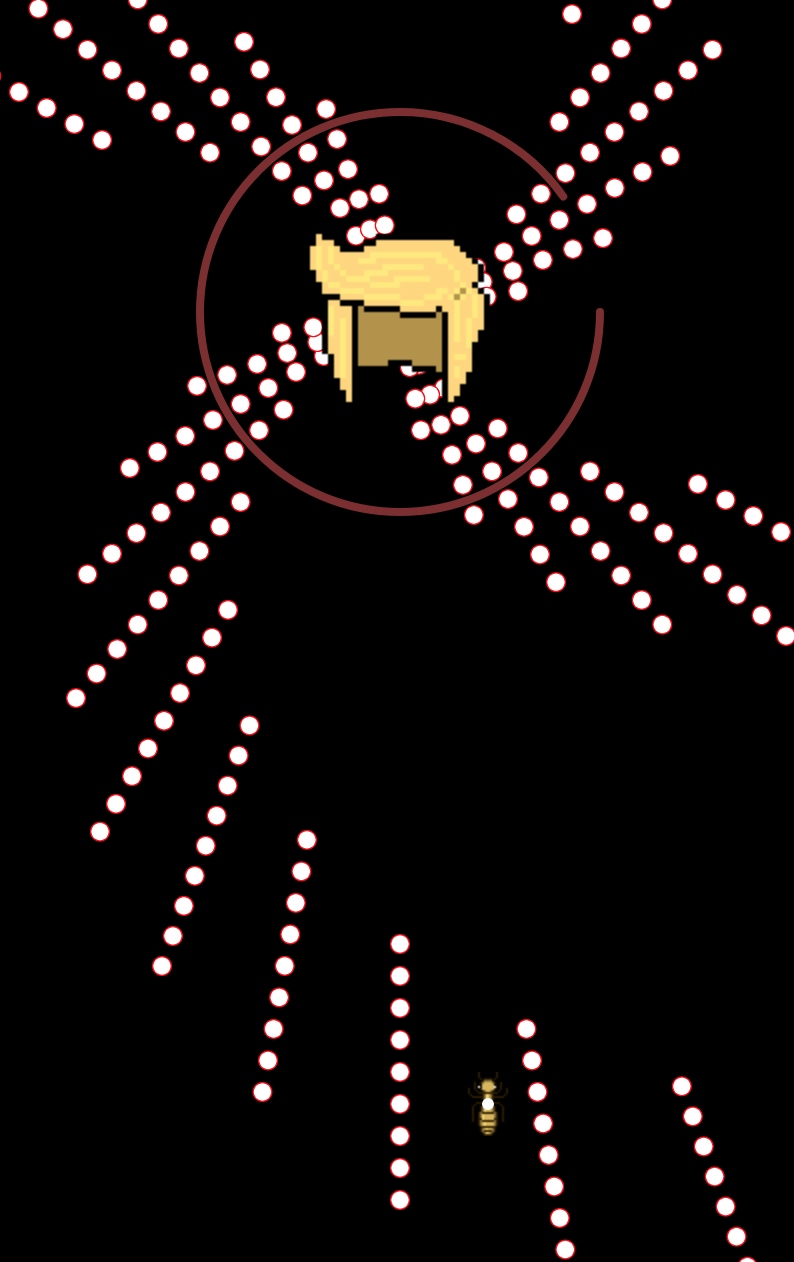}
  \caption{}
  \label{fig:spawnerOneTwo}
\end{subfigure}%
\begin{subfigure}{.5\linewidth}
  \centering
  \includegraphics[width=.9\linewidth]{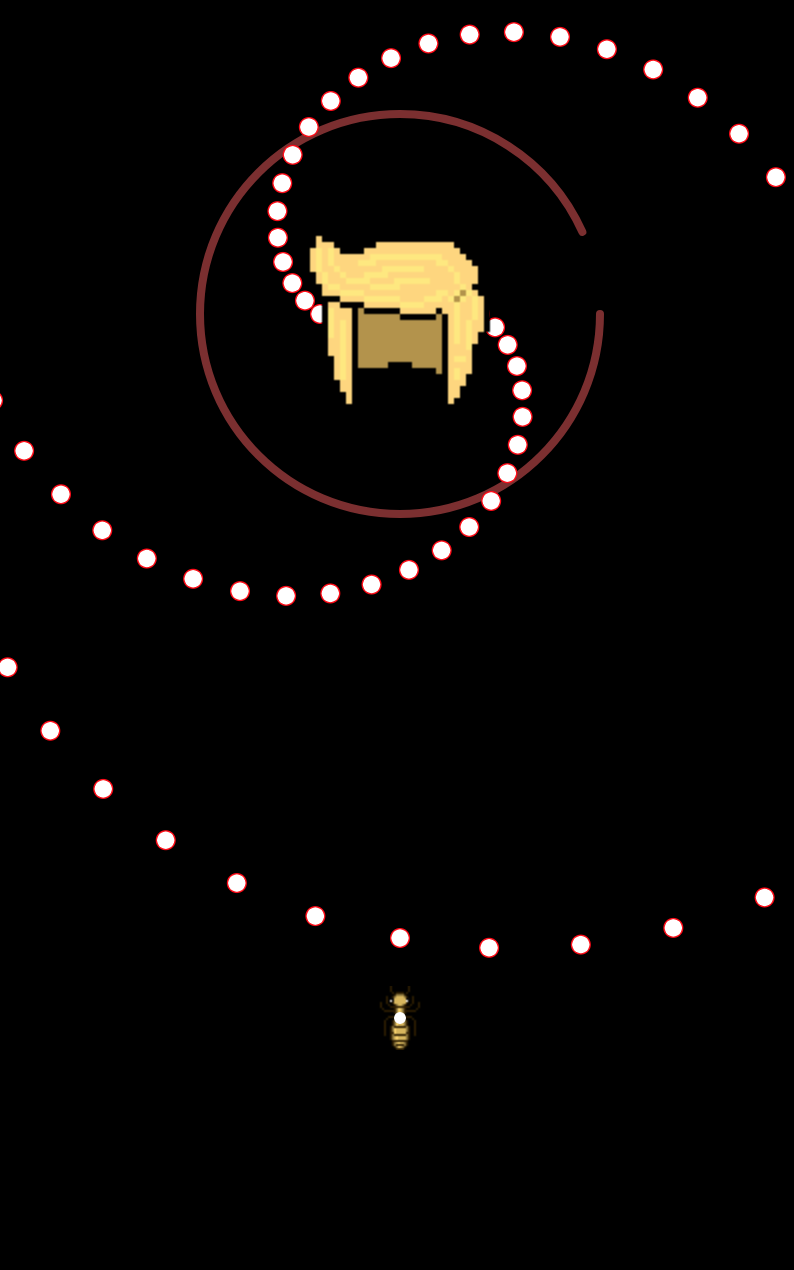}
  \caption{}
  \label{fig:spawnerThree}
\end{subfigure}
\caption{The visual representation of the spawners specified in Figure \ref{ex:talakatExample}. Figure \ref{fig:spawnerOneTwo} shows the output of spawners ``one'' and ``two''. Figure \ref{fig:spawnerThree} shows the output of spawner ``three''. The player is the small bug in the bottom of the screen, while the boss is the yellow wig from the top.}
\label{fig:spawnersExample}
\end{figure}

The spawners section contains information about the bullet spawners. Bullet spawners are invisible objects that are responsible for producing either bullets or additional spawners. The spawners section consists of an array of spawner definitions. Each spawner has a unique `id' to identify it, as well as parameters that define its spawning behavior. These parameters can include angle, speed, number of spawned objects, etc. Complex patterns can be generated by overlaying different spawners on top of one another. For example, Figure \ref{ex:talakatExample} utilizes three spawners: ``one'', ``two'', and ``three''. Spawner ``one'' generates 4 instances of spawner ``two'' evenly over an arc of $360^{\circ}$ (4 spawners at $90^{\circ}$ intervals) every 4 frames. Every 12 frames, the spawner rotates 10 degrees. Spawner ``two'' spawns 3 bullets evenly over an arc of $30^{\circ}$, each of which move at a speed of 4 pixels/frame. The end result of this pattern is a boss that fires 3 bullets in a small arc in four directions three times, rotates 10 degrees, and repeats. Spawner ``three'' spawns 2 bullets over an angle of $360^{\circ}$ while rotating 2 degrees per frame and changing direction once the spawner has rotated $180^{\circ}$, creating a pattern that fires bullets in a sweeping motion. Figure \ref{fig:spawnersExample} shows this spawner configuration in action.

\subsection{Boss Section}


The boss section contains information about the level. It defines boss health, boss position, and contains the level script which details boss behavior. Figure \ref{ex:talakatExample} contains an example of a simple boss section. Boss health controls the length of the level. In figure \ref{ex:talakatExample}, the length of the level is specified to be 3000 frames. For this version of Talakat, one point of boss health is depleted per frame regardless of player action, making health and duration one and the same. Boss position controls the placement of the boss in the level. In figure \ref{ex:talakatExample}, the boss will be in the upper center part of the level. The level script describes events that trigger when the boss' health reaches certain thresholds. In figure \ref{ex:talakatExample}, the boss has two events: the first event spawns spawner ``one'' and triggers when boss health is at 100\% (that is, the boss opens with this event), and the second event clears all of the previous spawners and spawns spawner ``three'' when the boss' health reaches 50\%.

\section{Constrained MAP-Elites}
Generating levels for a bullet hell game is a non-trivial problem. In addition to playability, one must consider difficulty, visual aesthetics, distribution, etc. Designing a fitness function to incorporate these dimensions is challenging. Additionally, there is more than one interesting level in the search space. For example: higher bullet count does not guarantee a more difficult level. Therefore, using a standard optimization algorithm or multi objective optimization algorithm may be insufficient as it may fail to consider possible optima in remote reaches of the search space. It is therefore important to use an algorithm that can comprehensively search multiple subspaces and identify the optima contained in those areas.

\subsection{MAP-Elites}
MAP-Elites~\cite{mouret2015illuminating} is a relatively new illumination algorithm that explores each area of the search space. Illuminating techniques are search algorithms that explore different areas of the search space and don't focus exclusively on the high performing areas. We can use the MAP-Elites algorithm to find playable levels that meet different criteria. For example, a level with very few bullets but still requiring high agent movement, a level with many bullets that cover a small area, and so on. An initial experiment using a vanilla implementation of the MAP-Elites algorithm failed to produce strong results. The recursive nature of Talakat scripts made it possible to generate millions of spawners in a fraction of a second. These spawners slowed down the generation process and made it difficult to explore the search space in a reasonable timeframe. In response, we decided to use a Feasible Infeasible 2 Population (FI2Pop) genetic algorithm~\cite{kimbrough2008feasible} on top of the MAP-Elites algorithm.

Constrained MAP-Elites is a hybrid algorithm that combines the illuminating functionality of MAP-Elites with the constraint-solving abilities of FI2Pop. As in standard MAP-Elites, this algorithm maintains a map of n-dimensions where each dimension is sampled. However, rather than a single chromosome, each cell stores two populations. One represents the feasible population which aims to maximize its fitness (measured as a function of playability), while the other is the infeasible population which attempts to satisfy a set of constraints. Every chromosome is located at a (cell, population) combination, and moves on those two levels: a cell level and a population level. A chromosome can move between cells if its properties change along one of the map's dimensions. A chromosome can also move between populations if it satisfies or fails to satisfy its feasibility constraints.

\subsection{Chromosome Placement}\label{Placement}

Upon generation, each chromosome is evaluated by an agent which is explained in section ~\ref{Agent}. The results of the evaluation determines the level's cell and population placements. The map used for the Constrained MAP-Elites has three dimensions: entropy, risk, and distribution. These dimensions were chosen based on the belief that they represent some aspect of difficulty in bullet hell levels. Entropy reflects the amount of input required of the player by calculating the information entropy using the first, second, and third derivatives of the agent's action sequence. In practice, entropy is correlated to number of times the player changed direction, stopped while moving, or began moving while stopped. Risk reflects the presence of bullets in close proximity to the player. It is calculated by dividing the screen into a grid and counting the number of squares around the player that contain bullets. Distribution represents the amount of space occupied by bullets. The distribution is calculated in a similar fashion to the risk, by dividing the screen into a grid and calculating the number of squares occupied by at least one bullet. These three values are calculated by the agent during its evaluation, and are used to place the level within its appropriate cell. 

During the time of evaluation, the agent is also determining the level's population placement. A chromosome is placed in the feasible population if it satisfies the following two constraints:

\begin{itemize}
\item Number of spawners doesn't exceed a fixed maximum value.
\item There are at least 10 bullets present for more than 50\% of frames.
\end{itemize}
The fitness of chromosomes in the feasible population is inversely proportional to the remaining boss health (the inverse of remaining boss health is an analog for survival time). A chromosome that fails these constraints is placed in the infeasible population, where fitness is calculated by multiplying the inverse of remaining boss health by the percentage of frames that contain bullets.

\subsection{Genetic Algorithm}
The populations within the individual MAP-Elites cells are evolved using grammatical evolution~\cite{o2003grammatical}. Grammatical evolution is distinguished from genetic programming algorithms in that grammatical evolution uses a mapping between a set of numbers and the grammar and acts on the numerical representation, whereas genetic programming algorithms operate directly on expressions. Each chromosome is represented as 11 arrays, each of which consists of 23 integers between 0 and 99. These numbers are mapped to the Talakat script using the grammar defined in figure \ref{ex:talakatGrammar}. The last array is used to represent level events. 
To create new chromosomes, we use uniform crossover over the 11 arrays, and uniform mutation on the integer values of one of the sequences. It is important to note that the parent chromosomes are retained, effectively implementing 100\% elitism. Parent chromosome selection occurs by first selecting a random cell from the map, and then using rank selection to select its parent chromosome. Once a new chromosome is generated, it is evaluated by the agent and placed in its cell and population as outlined in section ~\ref{Placement}. This paper will refer to a set of these mating events as a generation. If, after the creation of a generation, a cell's population exceeds its capacity, the lowest performing chromosomes are removed starting from the infeasible population. In short, the Constrained MAP-Elites algorithm operates as follows:
\begin{itemize}
\item Initialize, evaluate, and place the starting population within the appropriate cells and populations.
\item Run generation
\begin{itemize}
\item Create new chromosome
\begin{itemize}
\item Select parents
\begin{itemize}
\item Randomly select cell
\item Select chromosome using rank selection
\item Repeat if needed
\end{itemize}
\item Create new chromosome using crossover or mutation on selected parent(s) while retaining parent
\item Evaluate and place new chromosome in appropriate cell and population
\item Repeat until end of generation
\end{itemize}
\item Eliminate low performing chromosomes from cells that are at or above capacity
\end{itemize}
\item Repeat until end of experiment
\end{itemize}

\subsection{Agent}\label{Agent}
All chromosomes are tested using an A*~\cite{hart1968formal} agent. The A* agent's heuristic function consists of 4 parts: progress, lose, safety, and future location. Equation \ref{eq:AStarHeuristic} outlines the heuristic function used by the A* agent. $progress$ is the the number of frames the agent has survived so far, $lose$ is equal to 1 if the agent dies and 0 otherwise, $safety$ corresponds to the number of frames a completely stationary agent would survive at its current position, up to a maximum of 10, and $future$ corresponds to the agent's distance from point on the screen with the fewest surrounding bullets. $future$ is weighted less heavily as it is less important than agent survival, which is reflected by the other values, while $lose$ has the highest weight as the agent's survival is its highest priority.
\begin{equation}
f(x) = 0.5 \cdot progress - lose + 0.5 \cdot safety - 0.25 \cdot future
\label{eq:AStarHeuristic}
\end{equation}

Additionally, the agent's actions are constrained by setting two agent properties: dexterity error and strategy error~\cite{isaksen2017simulating}. Dexterity error forces the agent to repeat its actions for a number of frames. The severity of dexterity error is modeled as a gaussian distribution modeling the number of repeated frames. A high dexterity agent is forced to repeat fewer frames. The strategy error reduces the time alloted for the agent's decision making process. A high strategy error can force the agent to make decisions before it arrives at an optimal choice. A high strategy agent has more time to explore its options. By using different dexterity-strategy configurations for the evaluating agent, it is possible to guide the evolution toward patterns that heavily favor higher dexterity, higher strategy, or some combination of the two. For this work, dexterity and strategy each have 3 (low, medium, high) possible values, for a total of 9 possible dexterity-strategy configurations.

\section{Experiments}
We ran 9 experiments, one for each agent configuration. These experiments were run in parallel with the expectation that each experiment would independently generate interesting and playable levels tailored to the evaluating agent's dexterity and strategy level. Table \ref{tab:parameters} shows the strategy and dexterity values used during the experiment. The dexterity values are the standard deviation of the gaussian noise function representing the number of repeated frames. The strategy values correspond to the amount of decision-making time given to the agent in milliseconds for every frame.

In each experiment, we initialized the Constrained MAP-Elites with 100 random levels. The Constrained MAP-Elites uses crossover with 70\% probability and mutation with 30\% probability. Each map dimension is divided into 11 values (from 0 to 10). Each cell in the map has a population capacity of 50 chromosomes, shared between the feasible and infeasible sub-populations. A generation consists of 100 mating events. Each experiment was run for 24 hours.

\begin{table}
\begin{center}
\begin{tabular}{ |c||c|c| } 
 \hline
  & Dexterity & Strategy \\
 \hline
 \hline
 low & 10 & 40 \\
 \hline
 medium & 6 & 60 \\
 \hline
 high & 2 & 80 \\
 \hline
\end{tabular}
\end{center}
\caption{Values used for dexterity and strategy dimensions.}
\label{tab:parameters}
\end{table}

\begin{figure}
\centering
\includegraphics[width=\linewidth]{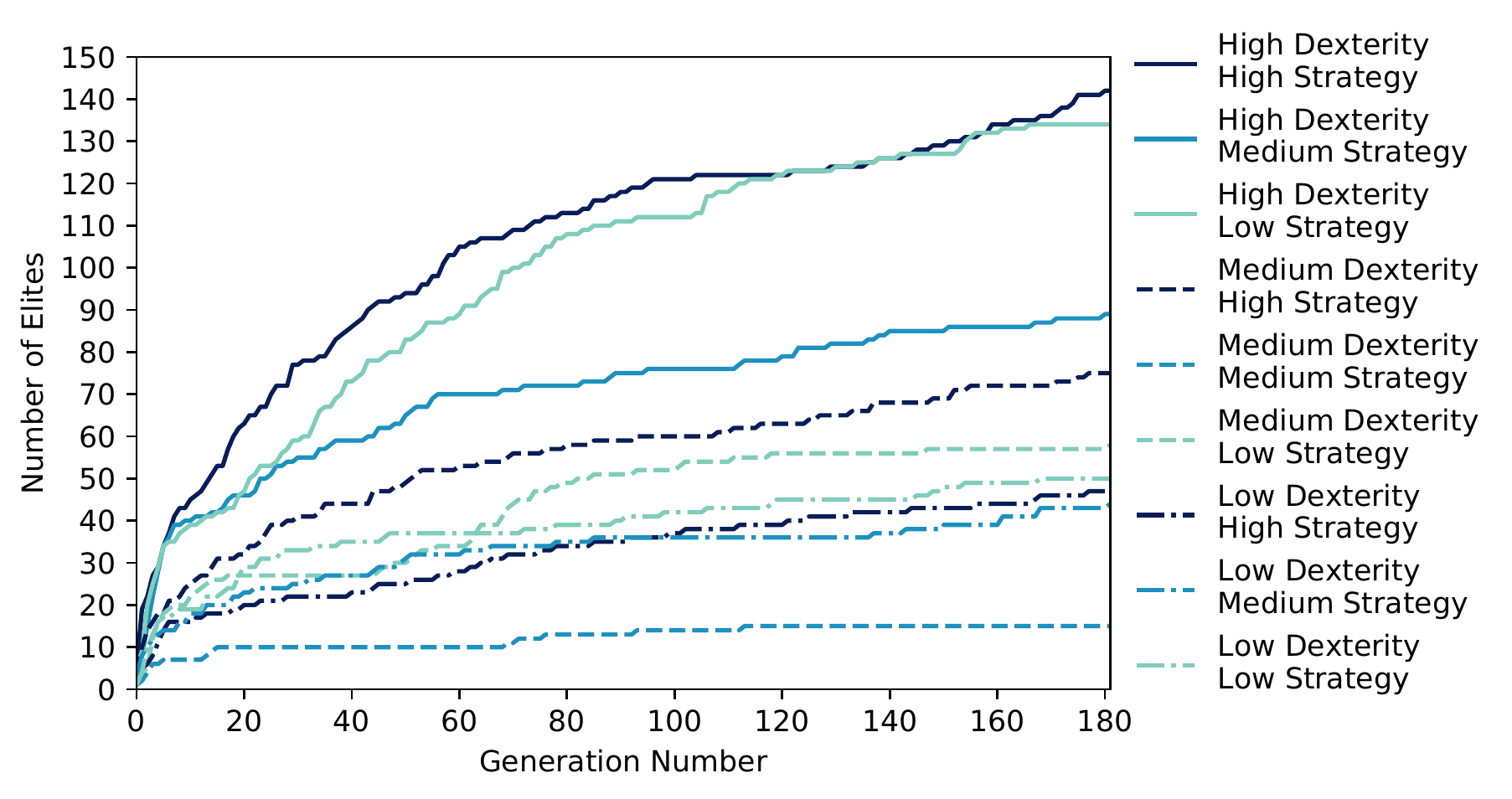}
\caption{Number of elites with fitness of 100\%.}
\label{fig:elitesNumber}
\end{figure}

\begin{figure*}
\begin{tabular}{ccc}
\begin{subfigure}{.3\textwidth}
  \centering
  \includegraphics[width=\textwidth]{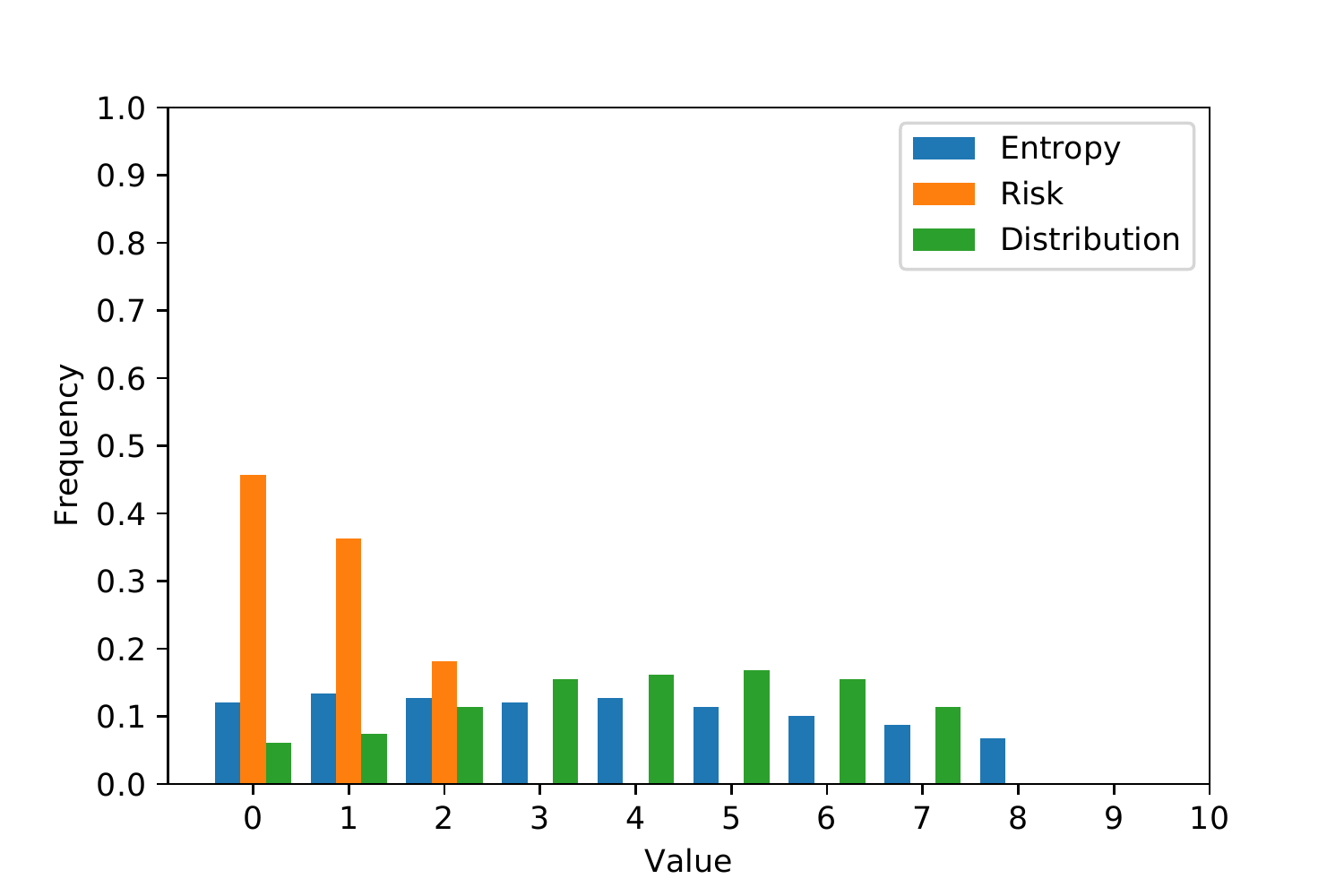}
  \caption{High Dexterity - Low Strategy}
  \label{fig:highLowHist}
\end{subfigure} &
\begin{subfigure}{.3\textwidth}
  \centering
  \includegraphics[width=\textwidth]{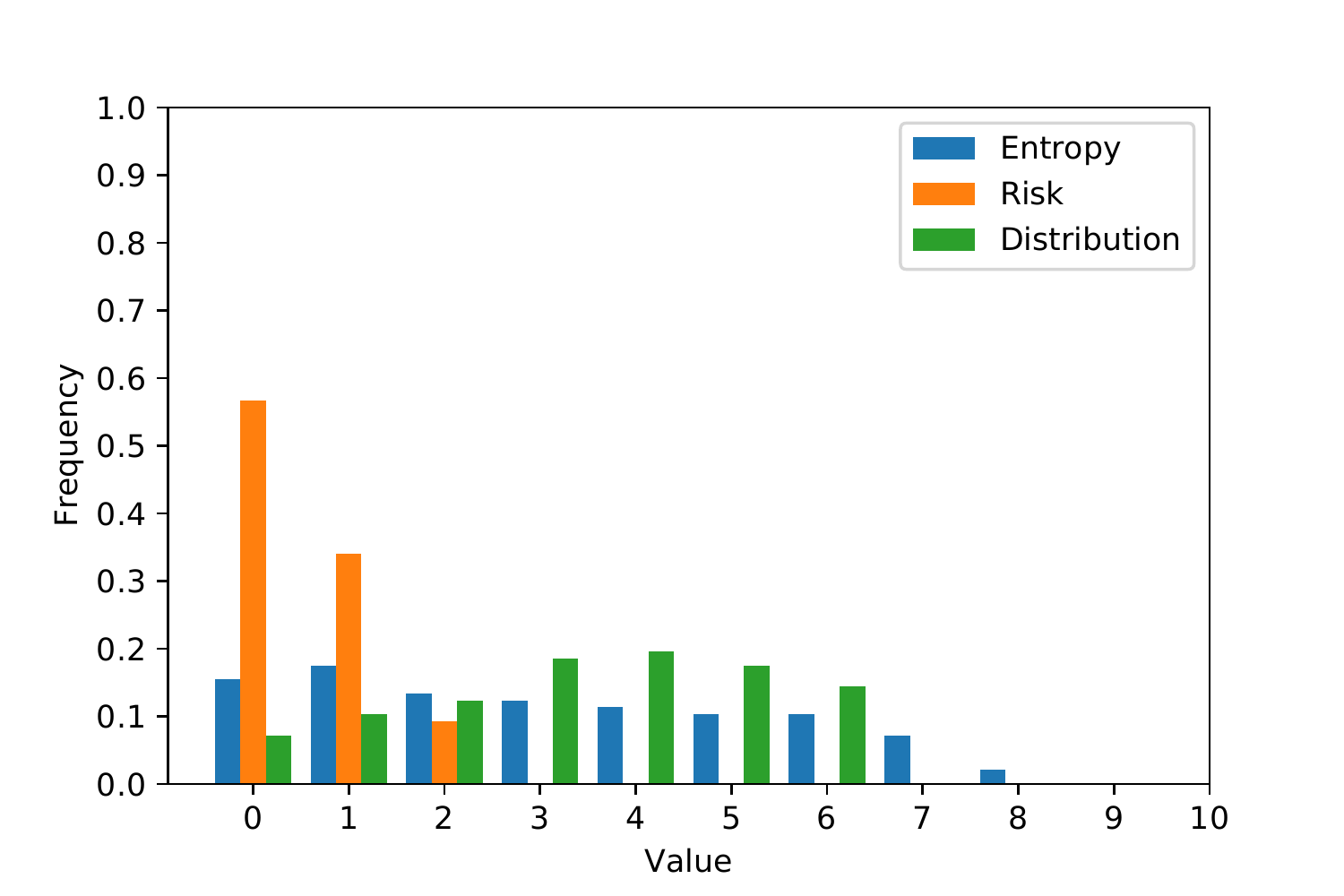}
  \caption{High Dexterity - Medium Strategy}
  \label{fig:highMedHist}
\end{subfigure} &
\begin{subfigure}{.3\textwidth}
  \centering
  \includegraphics[width=\textwidth]{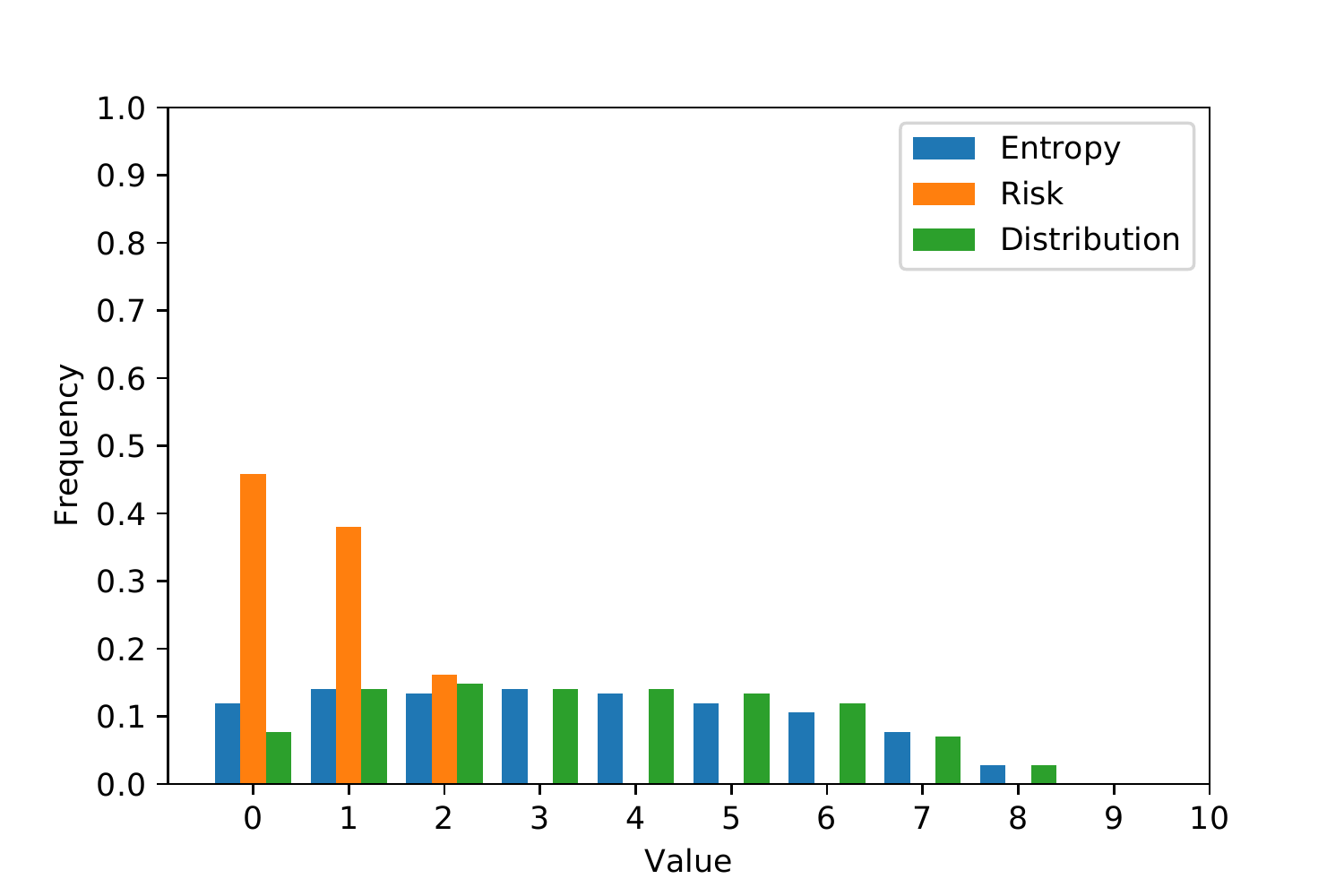}
  \caption{High Dexterity - High Strategy}
  \label{fig:highHighHist}
\end{subfigure}\\
\begin{subfigure}{.3\textwidth}
  \centering
  \includegraphics[width=\textwidth]{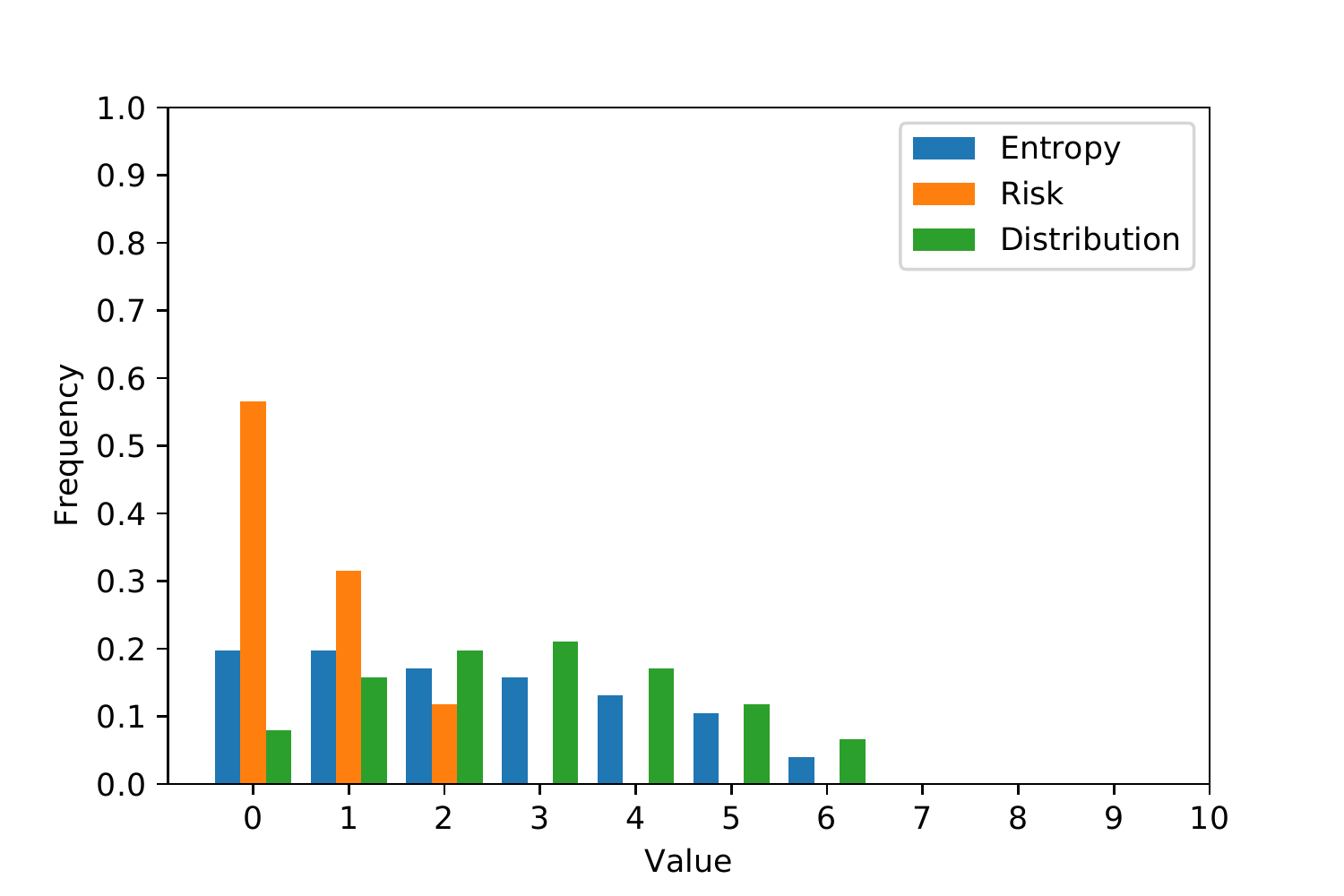}
  \caption{Medium Dexterity - Low Strategy}
  \label{fig:medLowHist}
\end{subfigure} &
\begin{subfigure}{.3\textwidth}
  \centering
  \includegraphics[width=\textwidth]{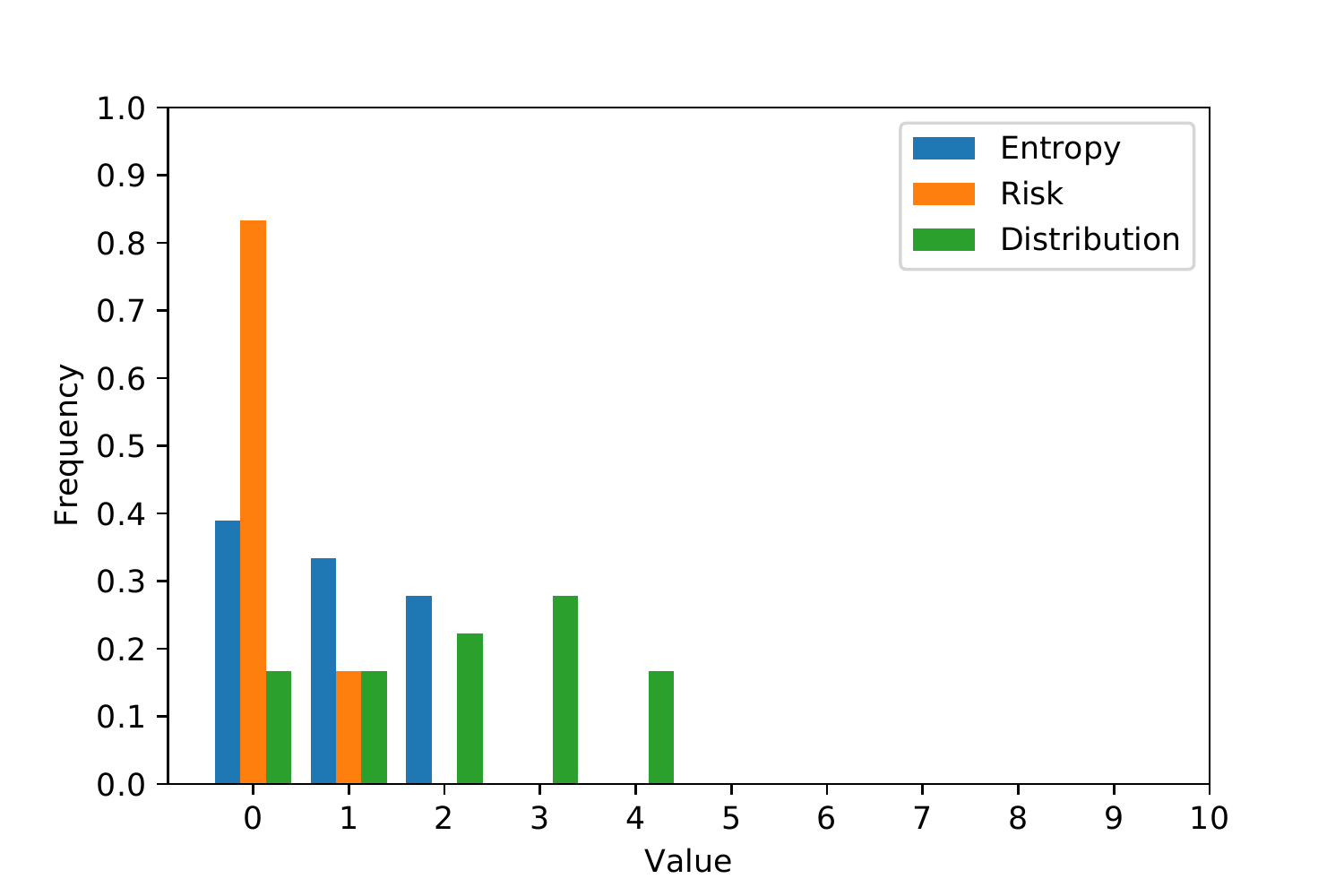}
  \caption{Medium Dexterity - Medium Strategy}
  \label{fig:medMedHist}
\end{subfigure} &
\begin{subfigure}{.3\textwidth}
  \centering
  \includegraphics[width=\textwidth]{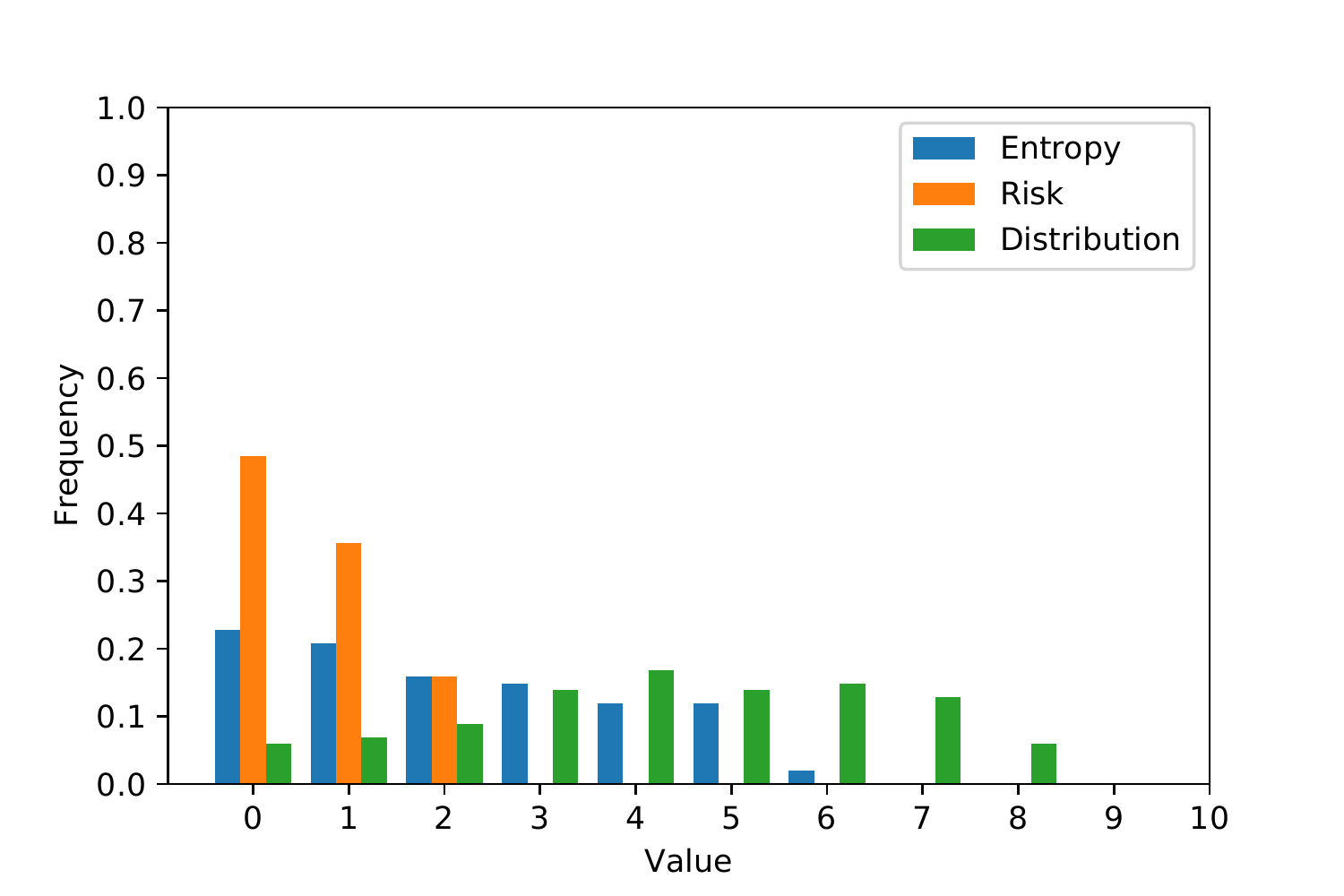}
  \caption{Medium Dexterity - High Strategy}
  \label{fig:medHighHist}
\end{subfigure}\\
\begin{subfigure}{.3\textwidth}
  \centering
  \includegraphics[width=\textwidth]{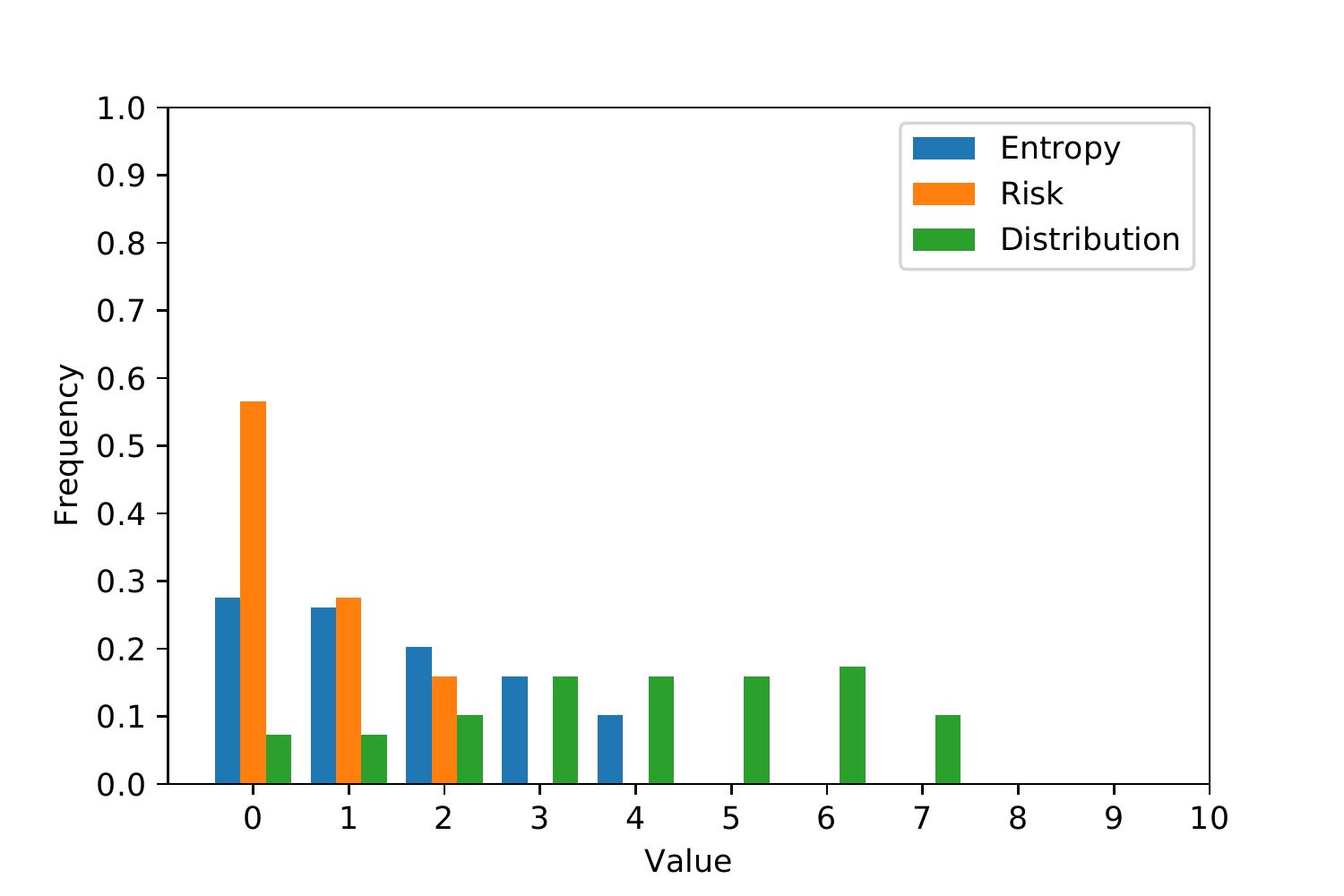}
  \caption{Low Dexterity - Low Strategy}
  \label{fig:lowLowHist}
\end{subfigure} &
\begin{subfigure}{.3\textwidth}
  \centering
  \includegraphics[width=\textwidth]{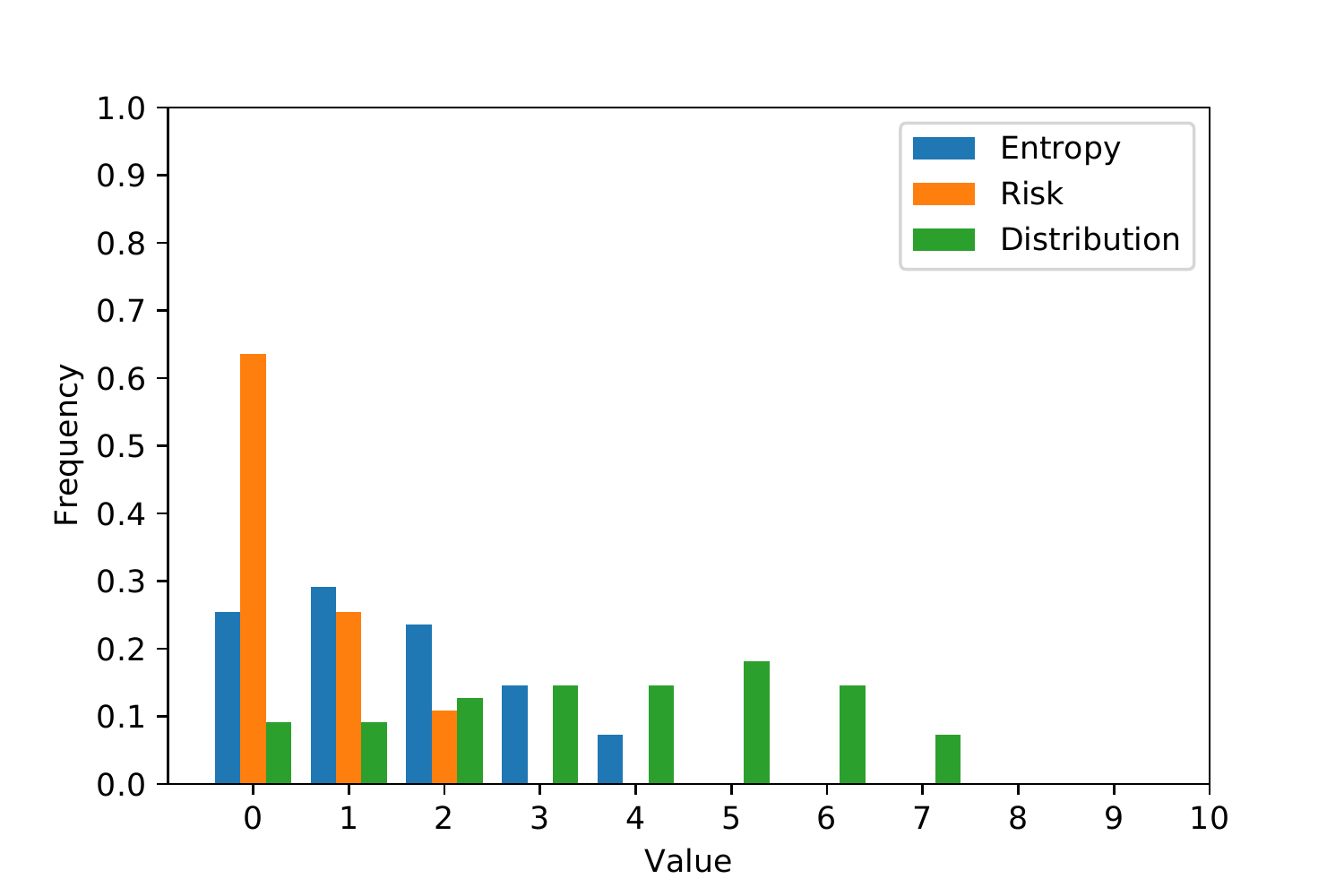}
  \caption{Low Dexterity - Medium Strategy}
  \label{fig:lowMedHist}
\end{subfigure} &
\begin{subfigure}{.3\textwidth}
  \centering
  \includegraphics[width=\textwidth]{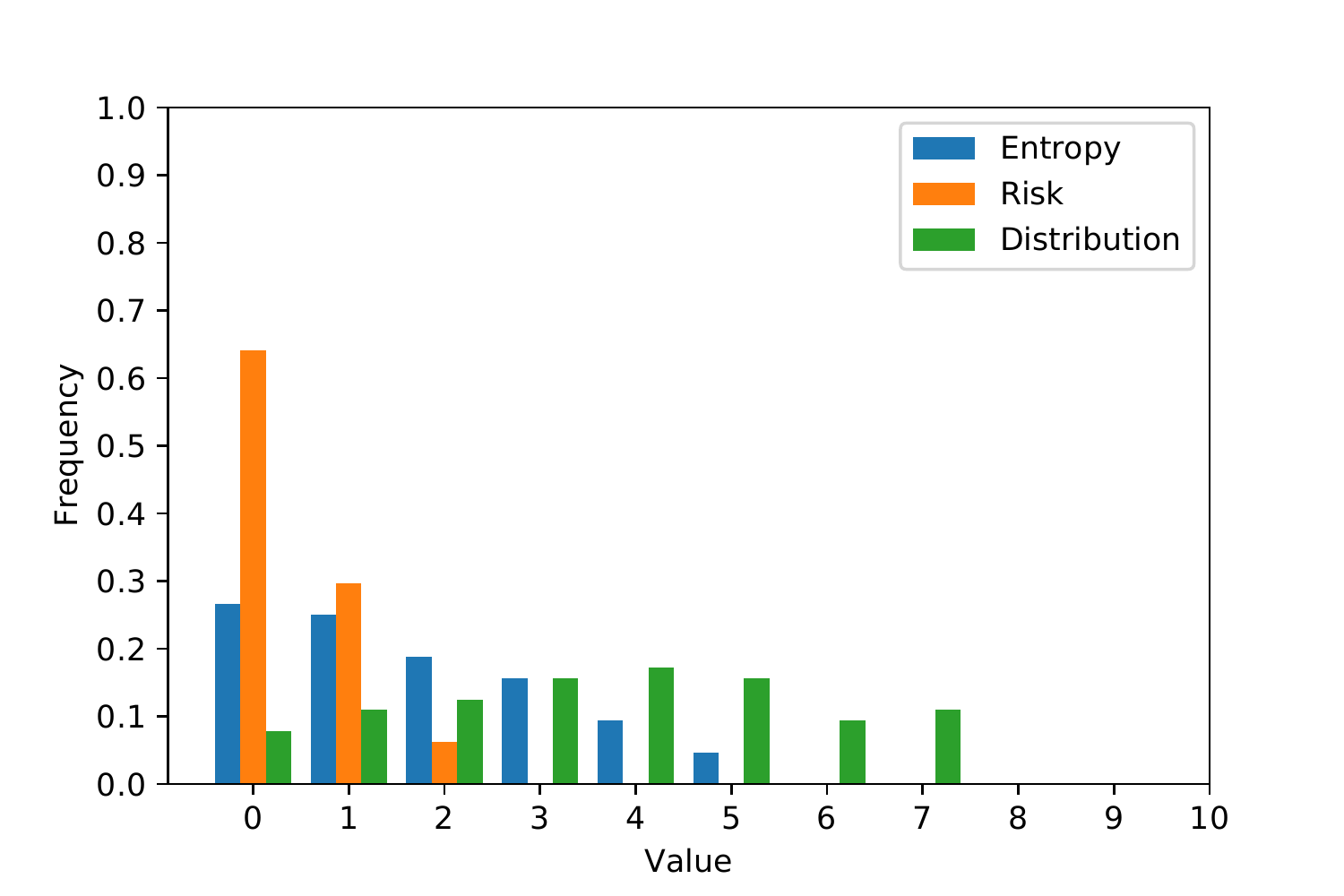}
  \caption{Low Dexterity - High Strategy}
  \label{fig:lowHighHist}
\end{subfigure}\\
\end{tabular}
\caption{Histograms for all the different 9 experiments after 24 hours. The x-axis represents the value of entropy, risk, or distribution depending on which bar one looks at. The y-axis represents the frequency with which a value with the corresponding value appears. That is, an orange bar at x=1 and y=0.4 would indicate that 40\% of the generated levels had a risk value of 1. The frequency values for a single evaluation dimension (entropy, risk, distribution) in a graph add up to 1. }
\label{fig:dimensionHistograms}
\end{figure*}


\section{Results}

We analyzed the number of elites generated with each generation for the constrained MAP-Elites for every configuration. Figure \ref{fig:elitesNumber} shows the number of elites with fitness of 100\% with every generation. The figure reflects how successful the constrained MAP-Elites has been at finding new elites with every generation. Because feasible levels are evaluated based solely on survivability, it is reasonable to conclude that the majority of the initial population was unsurvivable, with survivability increasing between generations.

Due to the nature of the agent parameters, the high dexterity - high strategy experiment went through fewer generations (approximately 180) than the low dexterity - low strategy experiment (approximately 1700). One generation of high dexterity - high strategy takes substantially more time than a low dexterity - low dexterity generation. This discrepancy is offset by the fact that high dexterity and high strategy experiments will find survivable levels in fewer generations, because the agents are better able to survive. It is also worth noting that the medium dexterity - medium strategy experiment has a highly anomalous graph. Upon investigation, this experiment was found to generate far fewer high-performing levels than the other experiments, which is especially surprising considering it occupies none of the extremes as far as dexterity and strategy parameters are concerned. It is possible that the population failed to mutate favorably over its run, but the exact cause is unknown.

Figure \ref{fig:dimensionHistograms} presents risk, entropy, and distribution (as discussed in section ~\ref{Placement}) of elite levels generated by each experiment. An elite level has a fitness of 1, the highest possible value, which indicates that the evaluating agent did not die when playing it. From figure \ref{fig:dimensionHistograms}, one can see that high dexterity experiments generally have more high entropy elite levels than low dexterity experiments. It seems obvious that high dexterity agents would be able to survive levels that would require more movement. However, this difference confirms that Constrained MAP-Elites is capable of generating a set of levels that are too demanding for the low dexterity agent, but are completable for the high dexterity agent, a set of levels which could be considered both non-trivial and playable. In this specific instance, any level generated by a high dexterity experiment with an entropy greater than 5 is will likely be too demanding for a low dexterity agent. There does not appear to be any noticeable difference in the three metrics with respect to strategy. It is possible that the difference between low and high strategy is too small to create any effect or that differences in strategy impact the levels in a way that cannot be expressed by entropy, risk, and distribution.

There are noticeable and notable flaws in the dimensions used by this implementation of Constrained MAP-Elites. For example, the distribution metric does not appear to be a good analog for difficulty. One would normally expect a negative correlation between a difficulty analog and the number of survivable levels, but this is not the case for distribution. In every one of the histograms, the frequency of each value of distribution appear to be fairly close, with no indication of a trend with respect to value. On the other hand, the risk metric exhibits this behavior quite strongly. There are far fewer survivable levels with a risk value of 3 than there are levels with a risk value of 1. However, the range of values for risk appear to be largely invariant across all experiments. It is entirely possible that risk is an effective difficulty analog that is unaffected by dexterity and strategy. This renders it less useful than entropy as a means of identifying more or less demanding levels in this specific experiment.

\begin{figure*}
\begin{tabular}{ccc}
\begin{subfigure}{.3\textwidth}
  \centering
  \includegraphics[height=0.28\textheight]{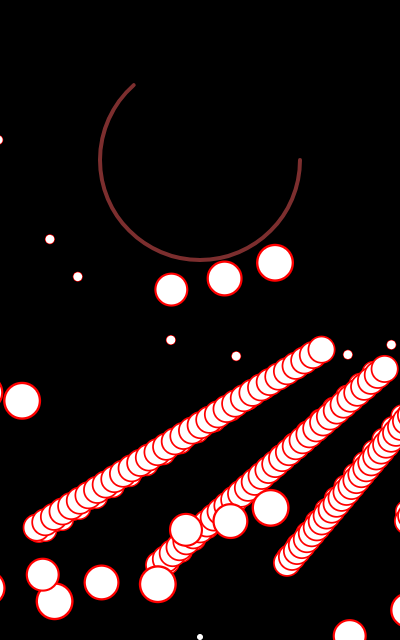}
  \caption{High Dexterity - Low Strategy}
  \label{fig:highLowBullets}
\end{subfigure} &
\begin{subfigure}{.3\textwidth}
  \centering
  \includegraphics[height=0.28\textheight]{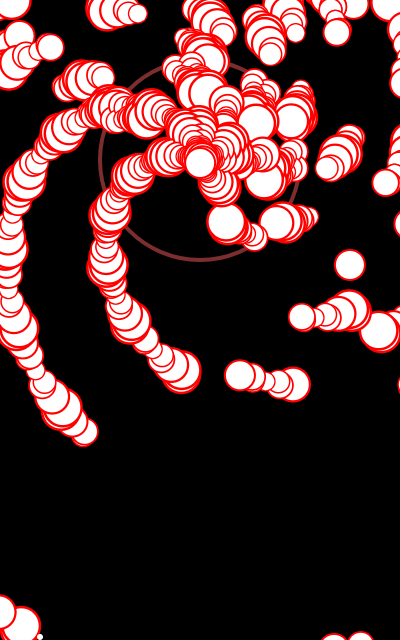}
  \caption{High Dexterity - Medium Strategy}
  \label{fig:highMedBullets}
\end{subfigure} &
\begin{subfigure}{.3\textwidth}
  \centering
  \includegraphics[height=0.28\textheight]{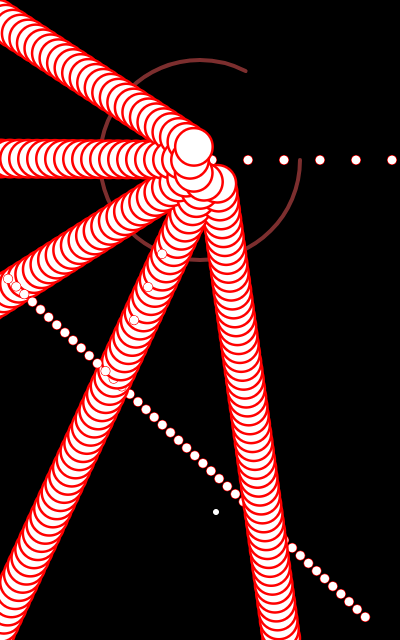}
  \caption{High Dexterity - High Strategy}
  \label{fig:highHighBullets}
\end{subfigure}\\
\begin{subfigure}{.3\textwidth}
  \centering
  \includegraphics[height=0.28\textheight]{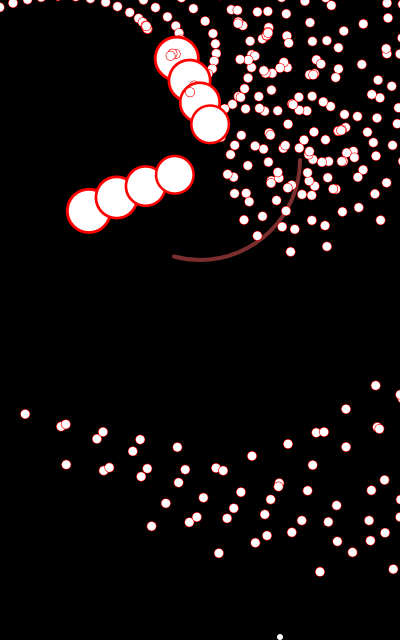}
  \caption{Medium Dexterity - Low Strategy}
  \label{fig:medLowBullets}
\end{subfigure} &
\begin{subfigure}{.3\textwidth}
  \centering
  \includegraphics[height=0.28\textheight]{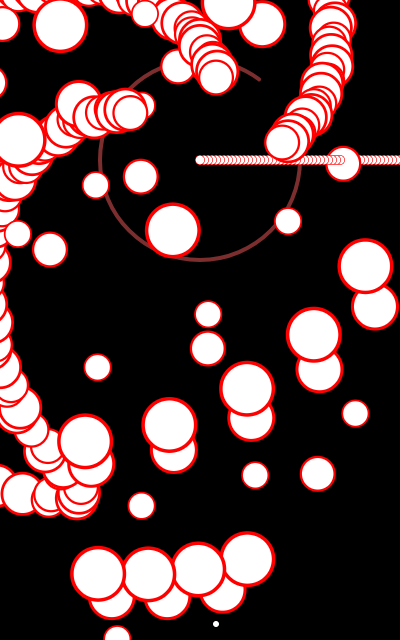}
  \caption{Medium Dexterity - Medium Strategy}
  \label{fig:medMedBullets}
\end{subfigure} &
\begin{subfigure}{.3\textwidth}
  \centering
  \includegraphics[height=0.28\textheight]{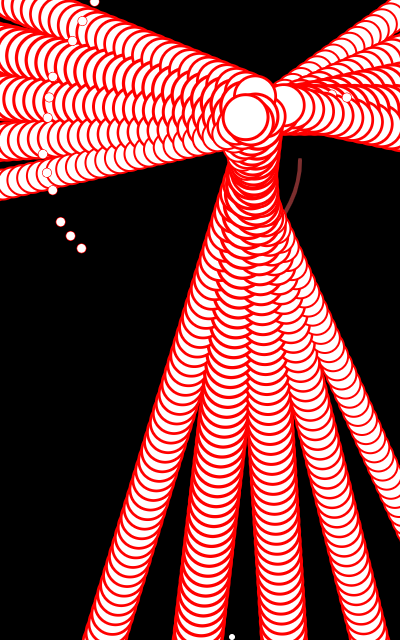}
  \caption{Medium Dexterity - High Strategy}
  \label{fig:medHighBullets}
\end{subfigure}\\
\begin{subfigure}{.3\textwidth}
  \centering
  \includegraphics[height=0.28\textheight]{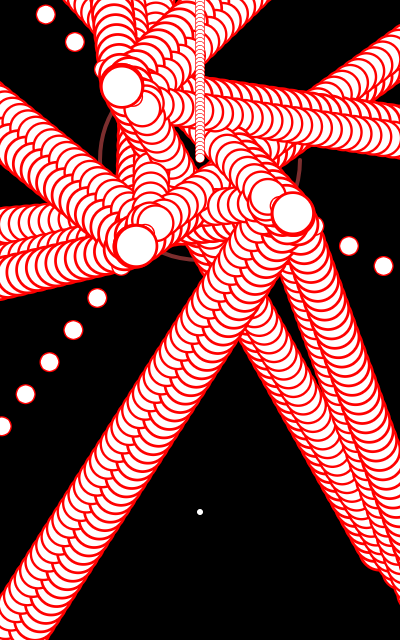}
  \caption{Low Dexterity - Low Strategy}
  \label{fig:lowLowBullets}
\end{subfigure} &
\begin{subfigure}{.3\textwidth}
  \centering
  \includegraphics[height=0.28\textheight]{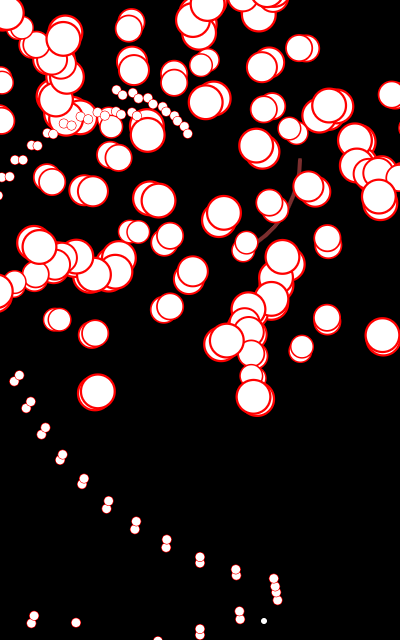}
  \caption{Low Dexterity - Medium Strategy}
  \label{fig:lowMedBullets}
\end{subfigure} &
\begin{subfigure}{.3\textwidth}
  \centering
  \includegraphics[height=0.28\textheight]{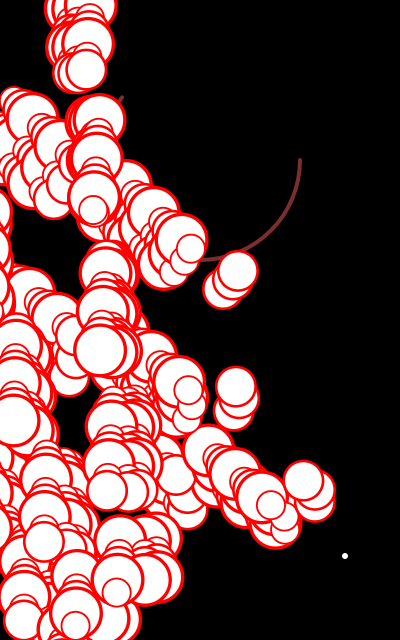}
  \caption{Low Dexterity - High Strategy}
  \label{fig:lowHighBullets}
\end{subfigure}\\
\end{tabular}
\caption{Examples of the generated levels for all the different 9 experiments. The high dexterity experiment images show levels with entropy 8, the medium dexterity experiment images show levels with entropy 6 (except image ~\ref{fig:medMedBullets} due to an anomaly), the low dexterity images show levels with entropy 4.}
\label{fig:dimensionBullets}
\end{figure*}

Figure \ref{fig:dimensionBullets} shows examples from the 9 experiments. We aimed to show levels with high entropy values to demonstrate the differences in difficulty generated by the experiments. The top row of images shows levels with entropy 8, the middle row shows images of levels with entropy 6, and the bottom row shows images of levels with entropy 4. Each of these values were chosen as they are higher than highest entropy achieved by an elite in an experiment with a lower dexterity. Therefore, levels in the top row should be too dexterously demanding for medium-dexterity agents, and levels depicted in the middle row should be too hard for low dexterity agents. Although the same relation cannot be definitively stated for strategy, visual observation and analysis shows distinctive differences in the amount of planning required for high strategy levels versus low strategy levels. For example, the level shown in figure~\ref{fig:lowHighBullets} begins with an empty stage, and quickly floods the left side with bullets after a short period of time. An agent without enough decision-making time to predict this will fail to move to the safe right side before it becomes closed off. Similar requirements are evident in the levels depicted by images ~\ref{fig:medHighBullets} and ~\ref{fig:highHighBullets}. Both levels open by splitting the level into sections, and firing bullets into certain sections some time later. A low strategy agent is less likely to be able to predict which sections will be safe in the time it is allotted, and inevitably die. This requirement is less pronounced but still present in medium strategy levels. Images ~\ref{fig:medMedBullets} and ~\ref{fig:medHighBullets} show levels with somewhat jagged walls of bullets. An agent can dodge the immediate threat by going between bullets, only to find itself trapped in the concave structure created by the jagged shape. From observing the images of levels created by the experiments, we believe that strategy did have some impact on the design of generated levels, even if the influence is not reflected in the statistics presented by figure ~\ref{fig:dimensionHistograms}.

\section{Conclusion}
In this paper, we presented Talakat a new framework that can be used to describe bullet hell levels. We also introduced a hybrid evolutionary algorithm called Constrained MAP-Elites that combines the MAP-Elites technique and the Feasible-Infeasible 2-Population genetic algorithm. We showed that the Constrained MAP-Elites can be used with Talakat to generate variations of levels. We suggest using Constrained MAP-Elites as a technique in level generation as game levels are very subjective. Instead of trying to define a ``good'' level, one can use multiple metrics as different dimensions of the Constrained MAP-Elites and utilize only playability for the fitness function. From the analysis of the histograms in the 9 experiments as well as high-performing levels, we confirmed that it is possible to create levels of varying difficulty.

For future work, we aim to investigate the possibility of generating aesthetically pleasing bullet hell levels in addition to challenging ones. A recent trend in bullet hell level design is the emphasis on patterns that are thematically coherent or pleasing to the eye. It may be possible to treat visual aesthetics as an additional dimension in the Constrained MAP-Elites or as part of a level's fitness function. How exactly that would be evaluated is unclear, although machine learning looks to be a promising avenue. We also aim to apply the Constrained MAP-Elites method to other types of games, for example those in the General Video Game Level Generation Framework~\cite{khalifa2016general}. We would also like to investigate the use of different metrics such as the ones presented by Liapis et al~\cite{liapis2013towards} as the dimensions for Constrained MAP-Elites.
\begin{acks}
The authors acknowledge the financial support from NSF grant (Award number 1717324 - "RI: Small: General Intelligence through Algorithm Invention and Selection.").
\end{acks}

\bibliographystyle{ACM-Reference-Format}

\end{document}